%% file: main.tex
\documentclass[preprint]{vgtc}               

\graphicspath{{figures/}{pictures/}{images/}{./}} 

\usepackage{times}                     

\usepackage{tabu}                      
\usepackage{booktabs}                  
\usepackage{lipsum}                    
\usepackage{mwe}                       
\usepackage{xcolor}
\usepackage{soul}
\newcommand{\add}[1]{\textcolor{black}{#1}}        

\usepackage{mathptmx}                  

\onlineid{1214}

\vgtccategory{Theory or Model}

\vgtcinsertpkg

\title{A Multimodal Reasoning Typology for \\ Grounding Chart-Image Coherence in Science Communication}

\author{
Avina Nakarmi\thanks{e-mail: an778@njit.edu}\\ %
     \parbox{1.8in}{\scriptsize \centering Department of Data Science \\ New Jersey Institute of Technology}
\and Sohom Sen\thanks{e-mail: ss4887@njit.edu}\\ %
     \parbox{1.8in}{\scriptsize \centering Department of Electrical \& Computer Engineering \\ New Jersey Institute of Technology}
\and Xun Song\thanks{e-mail: xs29@njit.edu}\\ %
     \parbox{1.8in}{\scriptsize \centering Department of Computer Science \\ New Jersey Institute of Technology}\\
\and Sreyashi Samaddar\thanks{e-mail:sreyashi.samaddar@brooklyn.cuny.edu}\\ %
     \parbox{1.8in}{\scriptsize \centering Department of Biology \\ Brooklyn College}
\and Aritra Dasgupta\thanks{e-mail: aritra.dasgupta@njit.edu}\\ %
     \parbox{1.8in}{\scriptsize \centering Department of Data Science \\ New Jersey Institute of Technology}
     }

\abstract{Charts and images appear together throughout scientific publications, yet most computational work does not characterize their coherence. We argue that a chart, its accompanying image, and the caption that links them form a multimodal unit, and that the inferential work required to read it varies systematically. To capture this variation, we develop a typology of reasoning gaps, R1 through R5, that characterizes how chart, image, and text jointly convey a scientific claim, and the interpretive work this demands of the reader. Some pairs restate the same data, while in other pairs, charts are used to quantify a structure the image localizes, project image content onto an external variable, audit an image-based claim, or jointly construct a frame that neither panel can establish alone. The typology is anchored in the grounding theory of communication and was derived bottom-up, with a neuroscience expert, from a corpus of 79 traumatic brain injury papers and \add{32} chart-image pairs. \add{Crucially, the levels provide a systematic mechanism for identifying where grounding succeeds or breaks down, rather than leaving it to subjective inference.} We show this in a study in which a domain expert and three non-experts judge \add{vision-language model (VLM)} descriptions of \add{25} pairs: \add{the level predicts where their judgments align and where they diverge, isolating the points at which contextual knowledge, not the figure, carries coherence.} \add{This typology thus offers figure designers a systematic way to balance text against chart-image pairs, bridging the expert-to-non-expert divide in reading a scientific takeaway.}}

\keywords{Multimodal reasoning, science communication, grounding, neuroscience, chart comprehension}

\begin{document}


\firstsection{Introduction}

\maketitle

\input{intro}

\input{related-work-v2}

\input{framework2}
\input{evaluation2}
\input{conclusion}

\bibliographystyle{abbrv-doi}

\bibliography{references2,references}
\end{document}

%% file: intro.tex
Charts and images often appear side by side in scientific papers. A connectivity matrix might sit next to a node-edge graph and a hierarchy tree of the same network (Figure~\ref{figs:RG}\textbf{a}); a study-design timeline might run alongside a year of pain-rating results (Figure~\ref{figs:RG}\textbf{e}); a brain map might be paired with an ROC curve showing how well a feature separates two groups (Figure~\ref{figs:RG}\textbf{d}). Often the pairing is the point: readers are meant to recover a claim from the combination that neither panel makes on its own.
This setting remains largely unmodeled in computational work. Much chart understanding work extracts data from one chart and answers questions about it~\cite{masry2022chartqa}. Multi-panel figure work tends to assume the relationships between panels are fixed or labeled in advance~\cite{siegel2016figureseer,kembhavi2016diagram}. \add{What goes unmodeled is the reader's side of the exchange: the inference needed to integrate panels when no question is posed and no relation is given. That inference is our subject, and what we formalize as interpretive effort.}
We treat the chart, image, and caption as a single multimodal unit for science communication. \add{The distinction between chart and image is functional, not visual: a \emph{chart} is a panel read through its quantitative encodings (a plot, matrix, or table), and an \emph{image} is a panel read as a depiction of structure (a scan, rendering, or micrograph). The same 3D rendering can function as either, depending on whether the reader is asked to measure it or to locate something in it.} By \emph{semantic interpretation} we mean the inference that connects what the panels show to the claim the figure makes; \emph{interpretive effort} is the cognitive work required for that inference, which our typology operationalizes as levels.

How much effort a pair demands varies with what the panels share. \add{When they share data, variables, or an explicit link, the reader has a shared footing;} in Figure~\ref{figs:RG}\textbf{a}, recognizing that two visuals encode the same network is nearly all that is asked. When they share none of these, as in Figure~\ref{figs:RG}\textbf{e}, the reader must bring the frame that binds them. What changes across this range is the amount of common ground the figure can assume. Clark's account of communication~\cite{clark1991grounding,clark1996using} makes this precise: an utterance is understood only against what the speaker and audience already share, and the effort to reach that footing rises as less of it is given. A chart-image pair works the same way, and the missing common ground corresponds to the interpretive effort the reader must supply.
We ground the typology in 79 traumatic brain injury (TBI) papers, working with a neuroscience expert whose figures routinely carry the claim in the pairing itself. We contribute a typology that classifies how chart, image, and text combine to convey a scientific claim; its bottom-up derivation with expert input; \add{and evidence of its diagnostic value: its levels predict where expert and non-expert readers agree on a figure's meaning and where they diverge.}

%% file: related-work-v2.tex
\section{Related Work}

\add{Our typology draws on three strands of prior work.}
\par \noindent \textbf{Semantic content of chart descriptions.} Lundgard and Satyanarayan~\cite{lundgard2022accessible} introduce a four-level model of natural-language descriptions of charts, running from elemental and encoded properties, through statistical and relational content and perceptual interpretation, to domain-grounded meaning. The model has since been used to evaluate machine-generated descriptions and to study accessibility for blind and low-vision readers~\cite{islam2024lvlmcharts,zong2022richdata}. \add{Their model asks what a single chart description should contain, and we borrow it directly as our instrument for measuring how much inferential work a description performs. Our own question sits one level up, at the inferential structure a \emph{pair} of panels encodes rather than the content of one description.}

 \begin{figure}[t]
 \begin{center}
 \includegraphics[width=\columnwidth]{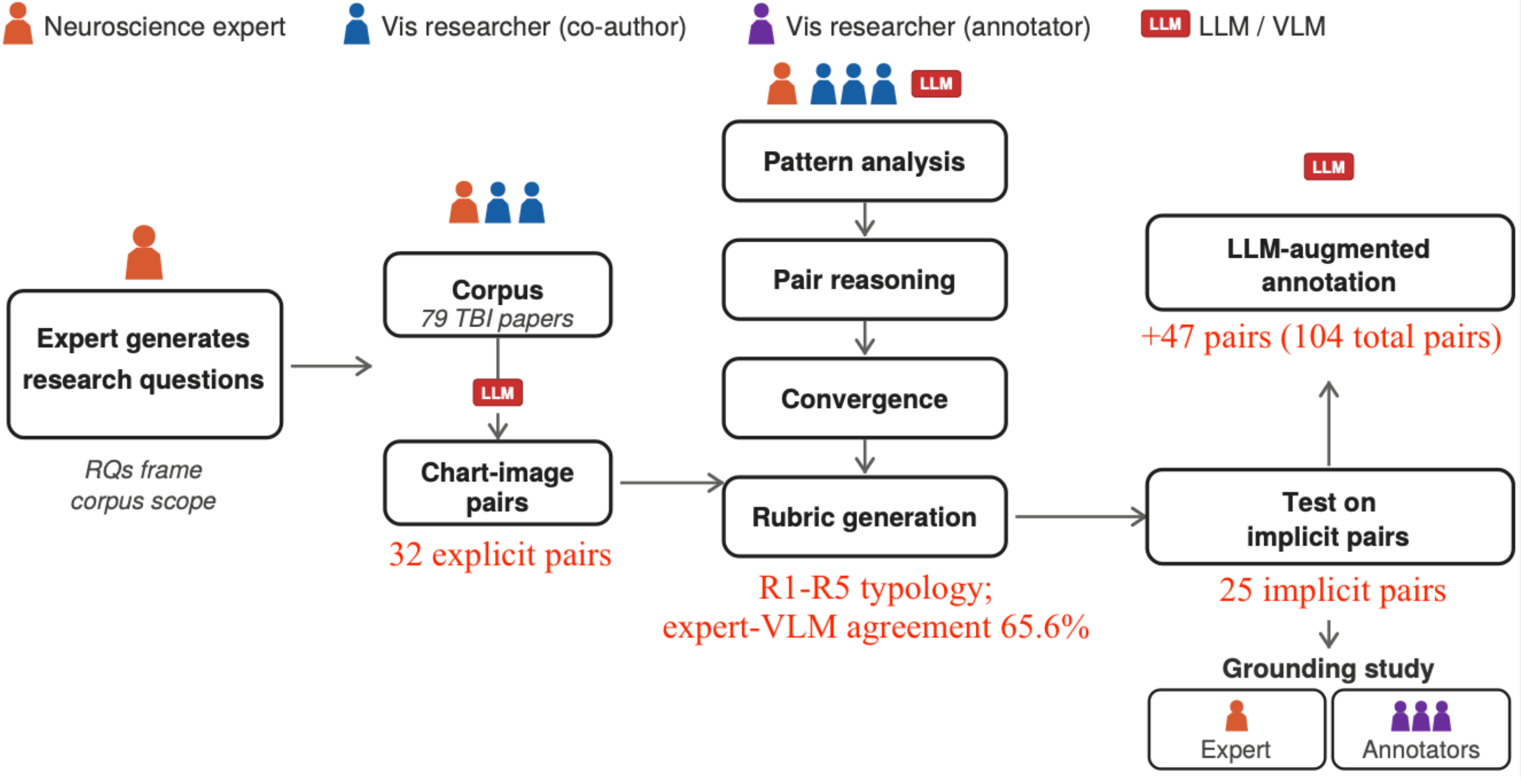}
 \vspace{-6mm}
 \caption{\label{figs: methodology}\textbf{Methodology pipeline.} The typology was derived bottom-up, not imposed. A neuroscience expert framed the corpus; three visualization-researcher co-authors iterated with her over explicit pairs to surface the R1–R5 levels; three separate annotators then tested them on implicit pairs. Successive rounds of prompt refinement ensured expert–VLM agreement is 65.6\%.}
 \end{center}
 \vspace{-9mm}
 \end{figure}

 \begin{figure*}[t]
 \begin{center}
 \includegraphics[width=.92\textwidth]{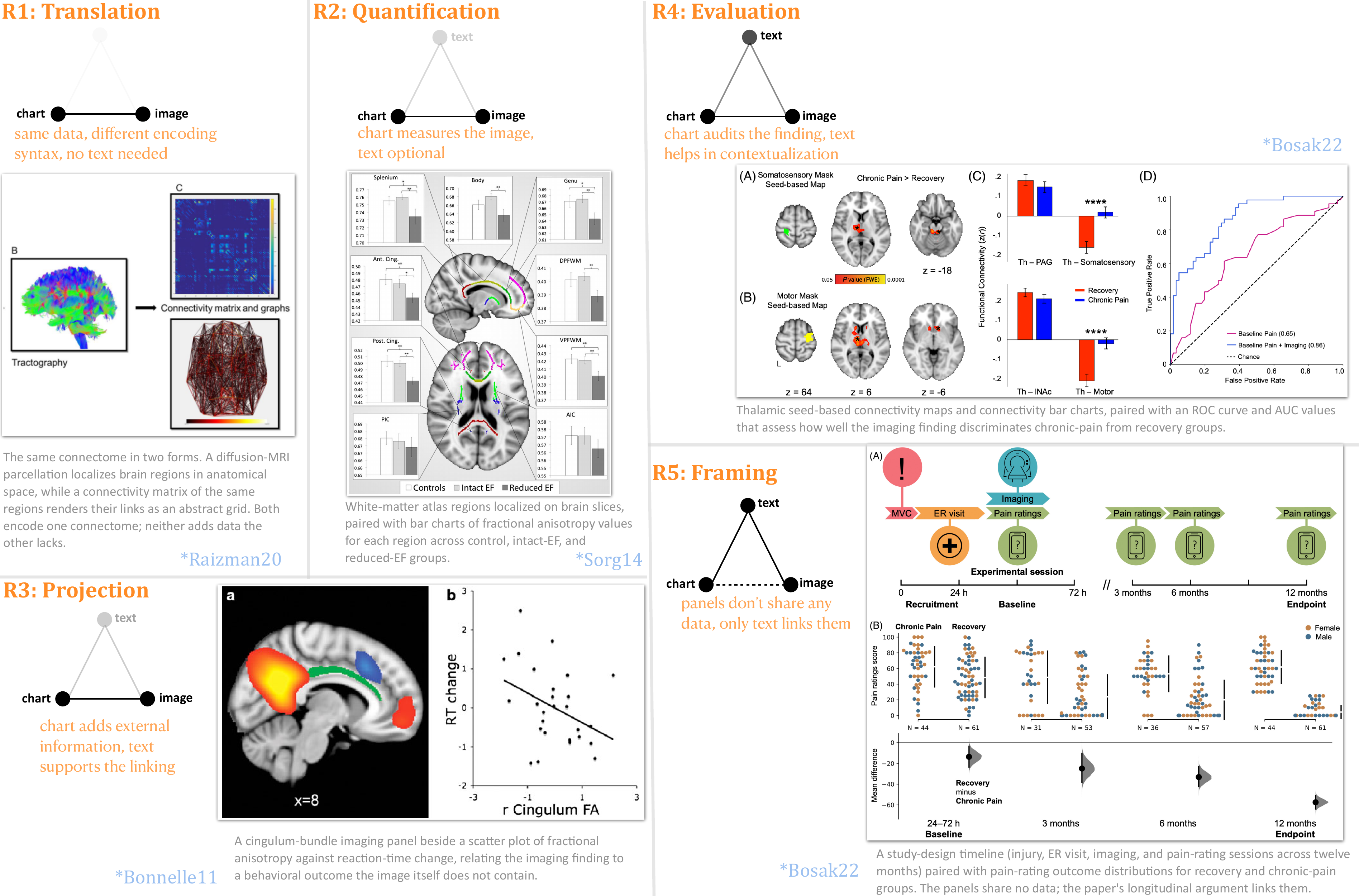}
 \vspace{-4mm}
 \caption{\label{figs:RG}\textbf{The R1–R5 typology illustrated with chart-image pairs from the TBI corpus}. Each panel shows a level's triangle icon~(solid vs dashed lines indicate the degree of interaction) and one corpus example. Interpretive effort grows from (a) to (e) as the link between chart and image moves outside the figure, and so does the contextual knowledge a reader must bring to ground it. At (a) R1 Translation~\cite{raizman2020traumatic} and (b) R2 Quantification~\cite{sorg2014white}, the panels share an object or measurement; expert and non-expert readers should ground the link similarly. At (c) R3 Projection~\cite{bonnelle2011default} and (d) R4 Evaluation~\cite{bosak2022brain}, the reader imports an external variable or validation criterion, opening room for divergent grounding. At (e) R5 Framing~\cite{bosak2022brain}, only readers who can interpret the contextual frame can ground the link between the chart and the image.}
 \end{center}
 \vspace{-9mm}
 \end{figure*}

\par \noindent \textbf{\add{Analytical reasoning and visual communication.}} \add{Reasoning has long been named as the goal of visual analytics, yet rarely formalized or operationalized systematically~\cite{dasguptaeurova}.} A long line of research establishes that a visualization is only as effective as the reader's ability to map what they see to the concepts the author intends~\cite{franconeri2021visualdata,kosslyn1989understanding}. \add{This literature treats the single visualization as the unit of analysis. Related work on visualization rhetoric and reasoning~\cite{hullman2011visualization,vaidya2020knowing} examines how a single visualization advances an author's argument; we ask, instead, how chart and image panels jointly do so. A smaller line studies how text and visual content integrate in documents~\cite{kim2018facilitating,zhi2019linking}, and we extend that concern to the scientific figure, where the caption is not auxiliary metadata but part of the unit that carries the relational claim.} Clark and Brennan~\cite{clark1991grounding} and Clark~\cite{clark1996using} develop an account of communication in which understanding requires that contributions be accepted into common ground, \add{with the effort varying by what participants already share.} Clark's primary case was face-to-face dialogue, but related notions of audience design and conversational implicature~\cite{clark1982hearers,grice1975logic} extend the account to other modes. \add{We build on this: grounding is the spine of our typology, and the reader's missing common ground is what each level measures.}
\par \noindent \textbf{Multimodal scientific figure understanding.} Recent machine learning research has tackled scientific figure parsing~\cite{siegel2016figureseer}, multi-panel interpretation~\cite{kembhavi2016diagram}, chart question answering~\cite{masry2022chartqa,kahou2018figureqa}, and benchmarks for vision-language models on scientific figures~\cite{wang2024charxiv,li2024mmsci}\add{, surveyed recently by Huang et al.~\cite{huang2024pixels}}. \add{These systems treat panels as content to extract or queries to answer, and reason within a chart rather than across a chart and its image. That cross-panel step is the clear gap: our typology gives a principled basis for testing whether such systems, with or without domain expertise, can recover the inferential structure a scientific figure builds across paired panels, whether explicit (a multi-panel figure) or implicit (figures linked through the paper text).}

%% file: framework2.tex
\section{A Typology of Multimodal Reasoning Gaps}
\label{sec:framework}

A chart-image pair invites semantic interpretation: recognizing what the panels share, aligning measurement with structure, or importing context neither panel supplies. This interpretive effort varies systematically, and we characterize the variation as a typology of reasoning gaps, R1 through R5, where each level names what the reader must supply to accept the pair's relational claim. The typology rests on Clark's account of grounding~\cite{clark1991grounding,clark1996using}, in which understanding requires contributions to enter common ground, at an effort set by what is already shared. We derived it bottom-up with a neuroscience expert from a corpus of traumatic brain injury (TBI) papers, fitting the level boundaries to the reasoning she performed when reading pairs. Across R1 to R5, the interpretive effort shifts from the figure to the reader, a progression we test as a falsifiable prediction in our grounding study.

\subsection{Expert-guided Corpus Selection}
The \add{typology} emerged from sustained collaboration\add{ through} a six-month, multi-step methodology~(Figure~\ref{figs: methodology}) with a neuroscience expert specializing in TBI \add{with} more than 10 years of experience. Initial discussions centered on a question we did not have a settled answer to: when she reads a paper, what does she do to connect the chart panel to the image panel beside it? Her descriptions surfaced a recurring distinction\add{: some pairs she resolved at a glance, while others required deliberate reasoning}. \add{Relating a spatial finding to a clinical scale, judging whether a chart audits an image's claim, holding two study systems together to read a longitudinal argument: these distinctions became the starting point for deriving the typology.}
We assembled a corpus of 79 TBI papers that addressed expert-generated research questions about symptoms, treatment, evaluation, and TBI outcomes\add{, collected by }searching for keywords following the expert's suggestions. Within each paper, we identified candidate chart-image pairs by inspecting every figure. The unit of analysis was the chart-image pair, with the surrounding caption and paper context used to resolve ambiguous panel relationships. A valid pair contained one chart panel (a quantitative plot, table, or matrix) and one image panel (an anatomical image, microscopy image, scan, or rendering), presented as a single figure or as adjacent sub-panels intended to be read together. \add{From this corpus we drew 104 chart-image pairs for annotation.}

\subsection{Iterative Typology Derivation}
\label{sec:iterative}
\add{We derived the typology through a structured cognitive walkthrough rather than free-form discussion or post-hoc coding. Working with the expert, three visualization-researcher co-authors read each chart-image pair and, for each pair, recorded the reasoning connecting the two panels using a fixed decision-point protocol. Each pair was routed through an ordered sequence of five decision points, from ``both panels show the same data'' through ``the chart evaluates the brain finding'' to ``named brain structures are the link,'' and the decision point that determined the classification was logged along with the assigned level. For every pair, the expert also recorded a confidence rating and whether the pair could plausibly belong to a second level. This last field made borderline cases explicit rather than forcing a single label: of 32 explicit pairs, 15 admitted an alternate classification, and 6 required inferential rather than direct reading. Convergence was assessed by examining these recorded decisions, holding that the triggering decision point was applied consistently across pairs without post-hoc reassignment, and that the expert and co-authors agreed on the relational claim each pair encoded. The decision points were instantiated for TBI figure conventions; the five levels are domain-general, but transferring the protocol would require re-specifying the decision points for another field.}
As an evaluation, we then ran a grounding study on \add{25} implicit pairs that do not appear together in the source paper but are linked via cross-referencing, with the expert and three non-experts (PhD students with visualization expertise but limited neuroscience background). Building the \add{typology} on explicit pairs and testing it on implicit ones lets us see how the reasoning gaps generalize to disparate panels connected only through text. To operationalize the \add{typology} at scale, we extend annotation to \add{104} pairs in total: \add{57 human-validated pairs (32 explicit from the derivation, 25 implicit from the grounding study) plus 47 additional pairs} annotated through an LLM-augmented workflow. \add{We verified a random 30\% of the LLM-assigned labels (14 of 47) against the typology, but did not audit the full set; the typology's claims rest on the 57 human-validated pairs, with the LLM-augmented set serving only to demonstrate the typology scales.}
The five levels are illustrated with corpus examples in Figure~\ref{figs:RG}. Each level tightens what the reader must supply: from recognizing shared content at R1 to constructing the cross-system frame at R5.
\par \noindent \textbf{R1 (Translation).}
The chart and image are alternative representations of the same underlying data. In Figure~\ref{figs:RG}\textbf{a}~\cite{raizman2020traumatic}, \add{a diffusion-MRI parcellation locates brain regions in anatomical space while a connectivity matrix and graph of those same regions render their links in abstract form}. The semantic interpretation here is recognition: the reader notices that the panels denote the same \add{connectome} under different visual grammars and accepts the equivalence. Interpretive effort is low, and the figure carries the linking work.
\par \noindent \textbf{R2 (Quantification).}
The chart provides scalar measurements of structures the image localizes. In Figure~\ref{figs:RG}\textbf{b}~\cite{sorg2014white}, white-matter atlas regions are paired with bar charts of fractional anisotropy values for each region. The reader aligns spatial reference with the measured value, taking on a step R1 does not require. The chart and image still share a common data substrate, so the link remains anchored in the figure. The interpretive effort is the alignment.
\par \noindent \textbf{R3 (Projection).}
The chart relates the image's content to a variable not present in the image. In Figure~\ref{figs:RG}\textbf{c}~\cite{bonnelle2011default}, a cingulum-bundle imaging panel sits beside a scatter plot of fractional anisotropy against reaction-time change, projecting the imaging finding into a behavioural outcome space. To accept the projection, the reader must import what the external variable means and what direction of relationship is being claimed. The caption and surrounding text often supply part of this material; contextual knowledge supplies the rest. Interpretive effort rises sharply.
\par \noindent \textbf{R4 (Evaluation).}
The chart audits the validity of the image's finding. In Figure~\ref{figs:RG}\textbf{d}~\cite{bosak2022brain}, thalamic seed maps are paired with connectivity bars and an ROC curve that subjects the imaging finding to discriminability assessment. The reader must hold an evaluative criterion in mind, understand what the chart is adjudicating, and accept the chart's verdict. The chart is not merely related to the image; it is being used to judge it. Contextual knowledge about validation procedures does much of the work, and the interpretive effort is correspondingly high.
\par \noindent \textbf{R5 (Framing).}
The chart and image jointly support an argument that neither panel can support alone. In Figure~\ref{figs:RG}\textbf{e}~\cite{bosak2022brain}, a study-design timeline runs alongside pain-rating outcome distributions across twelve months, asking the reader to accept a longitudinal argument about recovery and chronic-pain trajectories. There is no shared data between the panels. The link exists only because the reader brings to the page a frame in which the two panels belong together. Interpretive effort is highest here. The contextual frame, supplied by text and reader knowledge, does almost all the work.

\subsection{What the Reasoning Gaps Predict}
The R-levels track a progression: how much of the relational link between chart and image is carried by the figure, and how much must come from the caption, surrounding text, and the reader's background. As R increases, the figure does less of the linking work, and the reader does more. This yields a falsifiable prediction for systems with limited or uneven access to external material. Vision-language models are such systems, trained on broad corpora that supply some context but uneven technical depth in niche domains like TBI. The framework predicts that their performance should decrease as the reasoning gap increases.
\add{A learnability cross-check supports this prediction. Treating the expert's R-level assignments as a reference, we prompted a frontier vision-language model to label 32 explicit pairs across four iterative rounds, refining the prompt after each round from the disagreement pattern with the expert. Zero-shot agreement was low (11/32, 34\%), confirming that the levels cannot be recovered from the panels alone. Successive refinements raised agreement but plateaued; the largest gain came from adding the source paper's captions to the prompt (21/32, 65.6\%). This is not validation of the typology, which was derived through the process above; it is a downstream signal that the levels name something a context-poor system can only partially recover, and that the gain tracks the contextual material the framework says the reader needs.}

%% file: evaluation2.tex
\section{Preliminary Evaluation}

The reasoning typology predicts that as interpretive effort increases from R1 to R5, the relational claim between chart-image pairs might become increasingly dependent on contextual knowledge that the figure cannot fully supply. To test this, we ran a two-phase study of whether prior domain knowledge shapes how participants interpret VLM-generated descriptions of chart-image pairs, and whether that gap widens with the reasoning gap level.

\begin{figure}
    \centering
    \includegraphics[width=0.94\columnwidth]{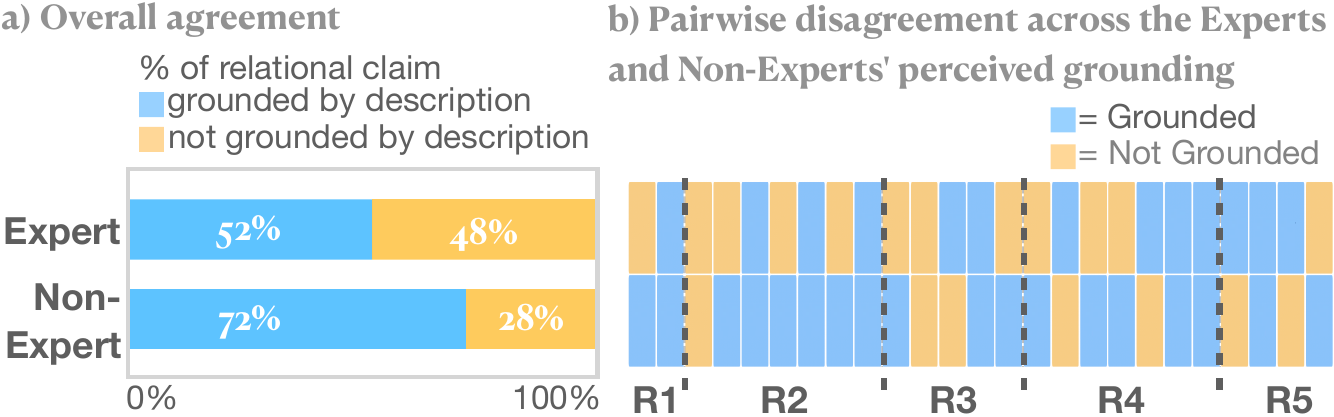}
        \vspace{-0.1in}
    \caption{\textbf{Disagreement in grounding of VLM-generated description.} Panel (a) shows that more non-experts considered the VLM capable of correctly describing the relational claim between chart-image pairs than the domain expert did, highlighting the impact of prior knowledge on filling reasoning gaps. Panel (b) shows agreement between experts and non-experts tends to decrease with the progression in reasoning gap levels~(from $50\%$ in R1, $57\%$ in R2, $40\%$ in R3, $29\%$ in R4 to $25\%$ in R5).}
    \label{fig:grounding}
    \vspace{-0.3in}
\end{figure}

Three doctoral students with CS/DS backgrounds and no prior TBI knowledge joined our expert collaborator as non-experts. Having taken a data visualization course and worked regularly with complex charts, they let us rule out visualization literacy as a confound: any interpretation difference traces to domain knowledge, not chart-reading ability.
We \add{drew the same 25 implicit pairs introduced in Section~\ref{sec:iterative}}, oversampling higher reasoning gaps, which are less visually perceptible and demand more effort, to test whether descriptions can bridge the harder cases. \add{A generator VLM (GPT-5.2), distinct from the learnability-check model, produced each description, prompted to explain the relationship without assuming prior knowledge and to describe each panel. These stood in for the figure's paper text, holding length and phrasing constant so that any expert--non-expert difference reflects reader knowledge, not source prose.}
\add{The two tasks were deliberately asymmetric, mirroring the roles in scientific communication.} Both ran in parallel, posing 3 questions per pair.
\par \noindent\textbf{Expert Assessment.} \add{As knowledge producer, the expert verifies whether a description transmits the intended claim.} She first saw only the pair and captions and recorded a takeaway surfacing a relational claim neither panel makes alone (a template was offered but her judgment took precedence); the description was withheld here to avoid bias. She then judged whether the description conveyed that claim and, \add{where it failed, why it captured only surface content and missed the ``why'' of the pairing}. The study across 25 pairs took roughly 2 hours.
\par \noindent\textbf{Non-Expert Assessment.} \add{As information consumers, non-experts decode the claim from the description alone.} They saw the pair with the description and judged whether it represented each panel and, most importantly, whether it explained the relationship. Each took 60--70 minutes.

\add{Our results show the R-level predicting where expert and non-expert judgments align and diverge, the diagnostic value the typology claims.}
\par \noindent{\textbf{Effect of Expertise.}}
Non-experts perceived grounding far more often than the expert (72\% vs.\ 52\% of pairs rated adequate; Figure~\ref{fig:grounding}a). The expert found that in nearly half of cases the description missed the relational claim she observed. \add{The expert's rejections rarely concerned vagueness; they turned on TBI-specific methods and claims that lay beyond what non-experts could catch.} Non-experts found most descriptions sufficient, a divergence that supports our hypothesis that prior knowledge supplies the context to fill reasoning gaps.


\par \noindent{\textbf{Systematic Disagreement Across Reasoning Gap Levels.}}
Disagreement between expert and non-experts increases as the level rises~(Figure~\ref{fig:grounding}b): at low levels the link is perceptible to both and they agree, while at higher levels the expert judges the claim inferentially from domain knowledge as non-experts rely on the VLM framing, and the two part. \add{The direction reverses at R5, where the expert accepts thinner descriptions than non-experts do, her knowledge silently filling the gap the description leaves while non-experts, with no shared data to anchor on, cannot reconstruct the claim.}

\par \noindent{\textbf{Semantic Grounding Patterns in Reasoning Gap Level.}} This prompted a finer analysis of the descriptions, using Lundgard et al.'s four-level semantic model to operationalize grounding depth, grouping sentences into visually grounding (L1, L2) and inferential (L3, L4) to measure how much contextual linking each description performs~\cite{lundgard2022accessible}. Descriptions with more inferential content were likelier to be seen by non-experts as grounding the claim, suggesting inferential content does the linking work non-experts cannot supply, substituting for the domain knowledge they lack. The expert was not similarly swayed: her knowledge let her evaluate the claim independently of the description's inferential depth. The descriptions' semantic composition showed no consistent pattern across levels.
These results support the hypothesis that prior domain knowledge is decisive in bridging the reasoning gap. \add{More than confirming that experts and non-experts differ, they show the R-level predicting where that difference appears and, at R5, reverses}: where the description fell short, the expert's knowledge filled the gap, and where it seemed sufficient, non-experts could not see what was missing.

%% file: conclusion.tex
\section{Implications for Future Research}
We view chart-image coherence as reader-centered reasoning: as we move from R1 to R5, the interpretive effort shifts from what the figure shows to what the reader brings. Our grounding study reveals the cost of that shift. Experts and non-experts agree less as levels rise, and by R5 the pattern reverses, the expert's prior knowledge filling gaps non-experts cannot cross. This is not a flaw in any single description; it is inherent to combining text and image.

There are diverse implications of our work. Expecting vision-language descriptions of scientific figures to match a single ``ground-truth'' caption ignores the work readers actually do, and our benchmarks must evolve accordingly. Our \textbf{R} levels complement Lundgard \& Satyanarayan's~\cite {lundgard2022accessible} framework: L1–L4 characterize the depth of a chart description; R1–R5, the reach of the inference linking chart and image. Crossed as a grid, they show not only how deep a description goes but whether that depth meets the relational claim the pairing makes. Chart producers can adopt the same lens: where does the linking work sit, and does the text give readers what they need to generate the inference?
As a next step, we will test the typology beyond TBI and expand the grounding study, which needs more readers from more backgrounds. \add{The most intriguing question is whether reasoning probes that retrieve multimodal content~\cite{sen2026pointers} could let a VLM recognize, on its own, when a pair's reasoning gap is too wide for its description to close (e.g., at high R-levels, where the linking depends on context the description omits)}

\section{Acknowledgment}
This work is supported in part by the PROTECT project, awarded by the U.S. Department of Energy’s (DOE) Office of Cybersecurity, Energy Security, and Emergency Response (CESER) to Pacific Northwest National Laboratory (PNNL) through solicitation RC-40125b-2023; by the Collaborative Research, Innovation and Strategic Partnerships (CRISP) grant at NJIT, and by the NSF grant 2326195.

\section{Generative AI Usage}
During the preparation of this manuscript, the authors used Claude (Version 4.8, Anthropic) to improve the readability, clarity, and consistency of all sections. After using this service, the authors revised and validated the text for factual accuracy. The authors assume full responsibility for the final content of this publication.

\section{Supplemental Materials}

All supplemental materials can be downloaded via this link:
\url{https://tinyurl.com/yzyp35rc}